\documentclass[runningheads]{llncs}
\usepackage{graphicx}

\usepackage{tikz}
\usepackage{comment}
\usepackage{amsmath,amssymb} 
\usepackage{color}

\usepackage[accsupp]{axessibility}  

\usepackage{amsmath}
\usepackage{amssymb}
\usepackage{booktabs}
\usepackage{comment}
\usepackage{tabularx}
\usepackage{subcaption}
\usepackage[accsupp]{axessibility}  

\usepackage[breaklinks=true,bookmarks=false]{hyperref}

\begin{document}
\pagestyle{headings}
\mainmatter
\def\ECCVSubNumber{43}  

\title{A Hyperspectral and RGB Dataset for Building Façade Segmentation} 

\titlerunning{Abbreviated paper title}
%
\author{Nariman Habili\inst{1} \and
Ernest Kwan\inst{2}\and
Weihao Li\inst{1}\and
Christfried Webers\inst{1}\and
Jeremy Oorloff\inst{1}\and
Mohammad Ali Armin\inst{1}\and
Lars Petersson\inst{1}}
\authorrunning{N. Habili et al.}
%
\institute{CSIRO's Data61 (Imaging and Computer Vision Group), Canberra, Australia\\ 
\email{nariman.habili@csiro.au}\\
\url{https://data61.csiro.au/}\\ \and
Australian National University, Canberra, Australia
}
\maketitle

\begin{abstract}
Hyperspectral Imaging (HSI) provides detailed spectral information and has been utilised in many real-world applications. This work introduces an HSI dataset of building facades in a light industry environment with the aim of classifying different building materials in a scene. The dataset is called the Light Industrial Building HSI (LIB-HSI) dataset. This dataset consists of nine categories and 44 classes. In this study, we investigated deep learning based semantic segmentation algorithms on RGB and hyperspectral images to classify various building materials, such as timber, brick and concrete. Our dataset is publicly available at \href{https://doi.org/10.25919/3541-s396}{\textcolor{blue}{CSIRO data access portal}}.

\end{abstract}

\section{Introduction}
\begin{figure*}[!htbp]
\centering
\includegraphics[width=0.93\textwidth]{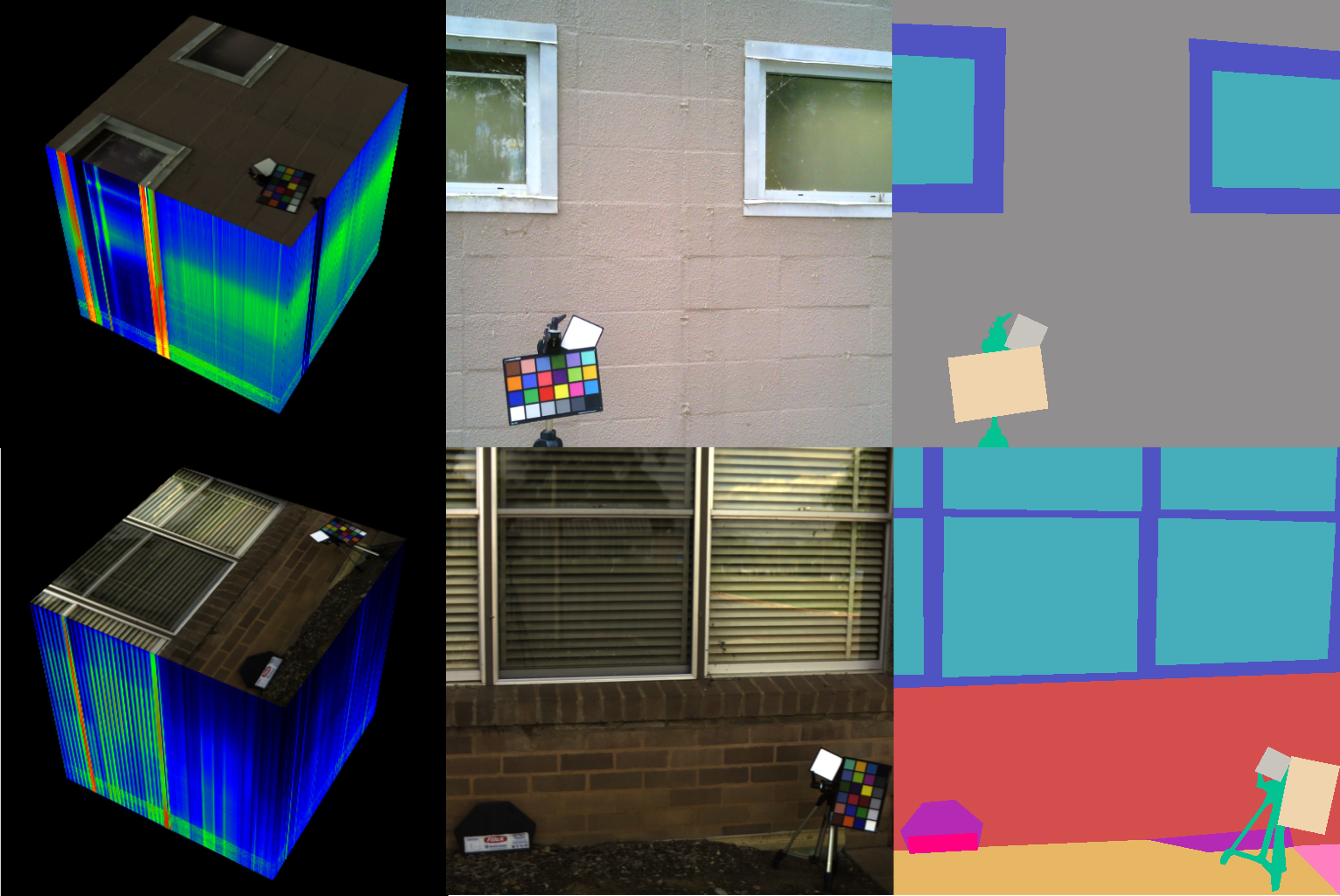}
\label{fig:test1}
\caption{Examples from our hyperspectral \& RGB façade dataset, the first column shows an hyperspectral image with 204 bands, the second column shows the corresponding RGB image and the last one shows the ground truth labels.}
\end{figure*}

Data on buildings and infrastructure are typically captured pre-disaster and post-disaster. In the case of pre-disaster inventory capture, the aim is to capture data on exposure, develop risk assessment models and inform governments and the private sector such as insurance companies about areas and infrastructure most vulnerable to disasters. On the other hand, a post-disaster survey is conducted to capture information about damage to buildings and other infrastructure. This information is used to assess the risk of future events, adapt and modify legacy structures and modify construction practices to minimise future losses. The information is also used to develop computational risk models or update existing models.
Pre-disaster and post-disaster surveys are usually conducted by foot, with surveyors visiting every building in a built environment and recording information about the make-up of the building in a hand-held device. Information such as building type, building material (roof, walls etc.) and the level of damage sustained by a building (in case of a post-disaster survey) are recorded. This is a laborious task and the motivation to automate this task formed the basis to collect data for this research project. Classifying building materials in a single image scene is called facade segmentation, i.e., the process of labelling each pixel in the image to a class, e.g., concrete, metal, and vegetation. 
While RGB images with three bands are commonly used in existing facade datasets such as ECP \cite{teboul2011shape}, eTRIMS \cite{korc-forstner-tr09-etrims}, Graz \cite{riemenschneider2012irregular}, and CMP \cite{tylevcek2013spatial}, hyperspectral imaging (HSI) divides the spectrum into many more bands and is rarely used for this purpose~\cite{dai2021residential}. In HSI, the recorded bands have fine wavelength resolutions and cover a wide range of wavelengths beyond the visible light. It provides significantly more information than RGB images and can be utilised in many real-world applications, such as agriculture, geosciences, astronomy and surveillance. 

The dataset presented in this paper was collected as part of a broader project to classify building materials for pre-disaster inventory collection. Images were taken of building facades in a light industrial environment. The dataset consists of 513 hyperspectral images and their corresponding RGB images, and each hyperspectral image is composed of 204 bands with a spatial resolution of 512 x 512 pixels. 44 classes of material/context were labelled across the images. Table \ref{tab:datasets} compares our LIB-HSI dataset with previous facade segmentation datasets.

\section{Related Work}

Hyperspectral imaging provides detailed spectral information by sampling the reflective portion of the electromagnetic spectrum. Instead of using only the three bands red, green and blue (RGB), the light emitted and reflected by an object is captured by hyperspectral sensors as a spectrum of hundreds of contiguous narrow band channels. The intensity of these bands are registered in a hyperspectral image. The human eye can only see reflections in the visible light spectrum, but a hyperspectral image can contain reflections in ultraviolet and infrared as well. By obtaining the spectrum for each pixel in the image of a scene, machine learning models can be trained to find objects and identify materials. It has been widely used in many real-world applications such as remote sensing \cite{ma2019deep}, food \cite{feng2012application}, agriculture \cite{dale2013hyperspectral}, forestry \cite{adao2017hyperspectral} and in the medical field \cite{lu2014medical}.

For each pixel in a hyperspectral image related to a material, there is a vector consisting of reflectance data. This vector is known as a spectral signature. These vectors form a hyperspectral data cube $(w, h, b)$ for an image with width $w$, height $h$, and $b$ bands. Whereas $b=3$ in the case of RGB images, $b$ can go up to hundreds in HSI. As different materials have their own spectral signatures, we can use this information to put pixels into groups. The hyperspectral image classification problem is the task of assigning a class label to each pixel vector based on its spectral or spectral-spatial properties. This is closely related to the semantic segmentation task in computer vision. Given an input image, the task is to output an image with meaningful and accurate segments. 

Despite the increasing interest in applying deep learning to hyperspectral imaging, the maturity is still low in this field. Compared to RGB images, the number of datasets available for HSI are significantly fewer and they are mostly old images of low resolution. Public datasets \cite{grana_veganzons_ayerdi} commonly used for benchmarking mostly consist of a single image that is captured by an Airborne Visible / Infrared Imaging Spectrometer (AVIRIS) sensor over some area. In addition, the boundary between segmentation and classification is unclear in HSI studies. Image classification usually refers to the task of giving one or more labels to the entire image, while image segmentation refers to assigning a class to each pixel in an image. A lot of research on HSI classification is actually doing semantic segmentation. For instance, state-of-the-art models in HSI classification\cite{roy2019hybridsn,roy2020attention,chakraborty2021spectralnet} performs segmentation by applying classification models on a pixel by pixel basis. In this study we also investigate the semantic segmentation algorithms on the LIB-HSI dataset. 

\begin{table}[!h]
\caption{Comparison of facade segmentation datasets}
  \label{tab:datasets}
\begin{tabular}{p{0.25\linewidth}p{0.25\linewidth}p{0.20\linewidth}p{0.30\linewidth}}
\toprule
datasets  & images   & classes  & modal                                                                                \\ \midrule
ECP \cite{teboul2011shape}   & 104  & 7 & RGB \\
eTRIMS \cite{korc-forstner-tr09-etrims} & 60 & 8 & RGB\\
Graz \cite{riemenschneider2012irregular}     & 50 & 4 & RGB \\
CMP \cite{tylevcek2013spatial}    &  606  & 12 & RGB \\
LIB-HSI(ours)   & 513 & 44 & RGB and hyperspectral   \\ 
\bottomrule                                                       
\end{tabular}
\end{table}

\section{Dataset}

The Specim IQ (Specim, Spectral Imaging Ltd.), a handheld hyperspectral camera, was used to acquire images for our LIB-HSI dataset. It is capable of capturing hyperspectral data, recovering the illumination spectrum of the scene (via a white reference panel or Spectralon) and deriving the reflectance cube as well as visualising the results. The wavelength range of the camera is 400 to 1000 nm. Each image consists of 204 bands and has a spatial resolution of 512 x 512 pixels. The hyperspectral images are saved in the ENVI format. In addition, the camera also converts the hyperspectral images to pseudo-RGB images, captures a separate RGB image (via its RGB sensor) and saves them all on a SD card.

Images for the LIB-HSI dataset were taken of building facades in a light industrial environment. For each building, multiple images from various locations and angles were captured. The characteristics of the illumination and sunlight are captured using a white reference panel. The derived reflectance image will contain errors if the lighting varies significantly across an image, e.g., due to shadows, as the Specim IQ camera will not be able to estimate the varying scene illumination with any accuracy. Therefore, all of the images were captured under shadow or overcast conditions. The LIB-HSI dataset consists of 513 images in total. The Scyven hyperspectral imaging software was used to visualize the images \cite{hab2015scyven}. The dataset was annotated by an external data annotation company. Every pixel was assigned a class label, making the annotated dataset suitable for supervised semantic segmentation. There are 44 classes in total, which are shown in table \ref{tab:classes}. Generally, they are grouped by their material class, some by their context. The goal is to correctly classify every pixel into one of the 44 classes and automate the identification of materials in scenes of buildings usually found in light industrial areas.

\begin{table}[h]
\begin{center}

\caption{Classes contained in the dataset}
  \label{tab:classes}
\begin{tabular}{p{0.1\linewidth}p{0.2\linewidth}p{0.57\linewidth}}
\toprule
Group & Superclass   & Classes Contained                                                                                \\ \midrule
1 &
  Miscellaneous &
  Whiteboard,   Chessboard, Tripod, Metal-Frame, Miscellaneous, Metal-Vent, Metal-Knob/Handle,   Plastic-Flyscreen, Plastic-Label, Metal-Label, Metal-Pipe, Metal-Smooth-Sheet,   Plastic-Pipe, Metal-Pole, Timber-Vent, Plastic-Vent, Wood-Ground \\
2      & Vegetation   & Vegetation-Plant,   Vegetation-Ground, Soil, Woodchip-Ground                                    \\
3      & Glass Window & Glass-Window,   Glass-Door, Concrete-Window-Sill                                                 \\
4      & Brick        & Brick-Wall,   Brick-Ground, Tiles-Ground                                                        \\
5 &
  Concrete &
  Concrete-Ground,   Concrete-Wall, Concrete-Footing, Concrete-Beam, Pebble-Concrete-Beam, Pebble-Concrete-Ground,   Pebble-Concrete-Wall \\
6      & Blocks       & Block-Wall                                                                                      \\
7      & Metal        & Metal-Sheet,   Metal-Profiled-Sheet                                                              \\
8      & Door         & Timber-Smooth-Door,   Door-Plastic, Metal-Smooth-Door, Metal-Profiled-Door, Timber-Profiled-Door \\
9      & Timber       & Timber-Wall,   Timber-Frame      \\ \bottomrule                                                       
\end{tabular}
\end{center}
\end{table}

There are some issues with the dataset that may make it unfavourable for training a deep learning model. Firstly, a dataset of 513 images with 44 classes is unusually small for training a neural network for semantic segmentation. Popular public RGB datasets have tens of thousands, or often more, images. 
To increase the size of the dataset, we performed data augmentation. Each image is cropped into nine patches, each of size 256 x 256 pixels. These patches are then rotated by $0^\circ$, $90^\circ$, $180^\circ$, and $270^\circ$. Hence, $4\times 9 = 36$ patches are created for every image. This resulted in over 15768 patches for training. For testing, the test set is kept in its original form.

In addition, the dataset suffers from a severe class imbalance. Many of the classes are only found in a few images and many only represent a small area of pixels. It will be very hard to recognize those classes compared to the majority classes for the network. How we choose to split the data set for training and testing needs to be carefully considered. Simply splitting it at random may lead to certain classes being completely omitted from training or testing. In order to have a roughly equal representation of classes in each set, we developed a new method to split the dataset. The proposed splitting method first preforms an analysis on the contents of the dataset (i.e., labels) and then makes the split according to the properties of the data, trying to retain the closest match between the profile of the data in the test and training sets. This can be achieved by the following procedure: (i) load each label image and create a vector descriptor of the image that is simply the number of pixels that belong to each class contained within the image; (ii) create an initial random 80/20 split of the data; (iii) calculate the average vector descriptor of each set; (iv) calculate a score for how well the sets align (using L2 norm distance between average vectors); (v) randomly swap images between the two sets and maintain the swap if the score is lower after the swap (i.e., a closer match between two sets); and (vi) repeat the previous step until a minimum is obtained.

\section{Experiments}

In this section, first, we briefly explain the evaluation metrics and the semantic segmentation algorithms applied to LIB-HSI dataset; second, the implementation details are explained.

\subsection{Metrics}
\label{sec:metrics}
We used the accuracy and the Intersection over Union (IoU) metric to evaluate the results of semantic segmentation on images. The accuracy is the fraction of pixels in the prediction having the same label as the corresponding pixel in the ground truth image. The IoU metric is defined for two sets, $A$ and $B$. It is calculated by the formula
\[ \text{IoU}(A,B) = \frac{|A \cap B|}{|A \cup B|}, \]
where $|X|$ denotes the size of a set $X$, and $\cap$ and $\cup$ are set intersection and union, respectively. Fixing an arbitrary order for all pixels in an image, the labels of all pixels in an image represent an ordered set. Thus two images can be compared using the IoU metric.
Then the IoU is lower thresholded to 0.5 and the result linearly mapped to the interval $[0, 1]$. Let $x$ be the IoU, the score $y$ is then given by
\[ y = 2 \max(x, 0.5) - 1 . \]

For metrics per class, precision, recall and IoU are used. Precision measures the proportion of positive predictions being correct, while recall measures the proportion of actual positives being correctly predicted. First, we find the number of true positive (TP), false positive (FP) and false negative (FN) labels for each image. Then we calculate the metrics with
\begin{align*}
    \text{Precision} &= \frac{TP}{TP+FP}\\
    \text{Recall} &= \frac{TP}{TP+FN}\\
    \text{IoU} &= \frac{TP}{TP+FP+FN}
\end{align*}

During training, the accuracy and IoU based score of the model are computed on every epoch. After that, we select the best performing model on the validation set and test it on the test set. We used the metrics above to evaluate how well a model performs on the test set.

\subsection{Segmentation Models}
Among well established semantic segmentation algorithms applied on hyperspectral images, we chose Fully Convolutional Network (FCN) \cite{long2015fully} and U-Net  \cite{ronneberger2015u} with the ResNet backbone \cite{he2016deep} in our experiments.  

\paragraph{FCN} is a deep neural network architecture used mainly for semantic segmentation \cite{long2015fully}. It does not employ fully connected layers and solely uses locally connected ones, such as convolution, pooling, and upsampling. Since FCN avoids using dense layers, it means fewer parameters and makes the networks faster to train. It also suggests that an FCN can work for any image size because all connections are local. A downsampling path is used to extract and interpret the contextual information, and an upsampling path is used to help the localization.

\paragraph{U-Net} is a CNN architecture developed for biomedical image segmentation \cite{ronneberger2015u}. The architecture is based on the FCN. U-Net consists of a contracting path and an expanding path. Hence it is a u-shaped architecture. The contracting path consists of a series of convolutional layers and activation functions, followed by max-pooling layers to capture features while reducing the spatial size. The expanding path is a symmetric path consisting of upsampling, convolutional layers, and concatenation with the corresponding feature map from the contracting path. An important aspect of U-Net is its ability to work with few training images, which is suited for HSI. In addition, we need an end-to-end network to output a segmentation map for an input image instead of CNNs used in HSI classification that output a single label for each input. U-Net is one of the most popular methods for image segmentation and is known to work for many different applications. In this work, we used a U-Net with the ResNet backbone~\cite{he2016deep}, which was equipped with Squeeze-and-Excitation (SE) blocks~\cite{hu2018squeeze}. 

\subsection{Objective Function}
To tackle the issues of the imbalanced dataset, we used focal loss~\cite{lin2017focal}. Unlike cross-entropy loss, these losses are designed to penalise wrong classifications more and focus on training the hard samples.

\subsection{Implementation}
All weights were randomly initialised and trained using the back-propagation algorithm. The Adam optimizer \cite{kingma2017adam} was used as it is popular and compares favorably to stochastic gradient descent (SGD) in practice. We used a learning rate of 0.001, but the ReduceLROnPlateau scheduler from PyTorch was used later to introduce a decay. The scheduler halves the learning rate when the validation loss stops improving. We used a batch size of 8 and trained the models for about 50 epochs.

\section{Results and Discussion}
\begin{figure}[!htb]
\centering
\includegraphics[width=0.95\linewidth]{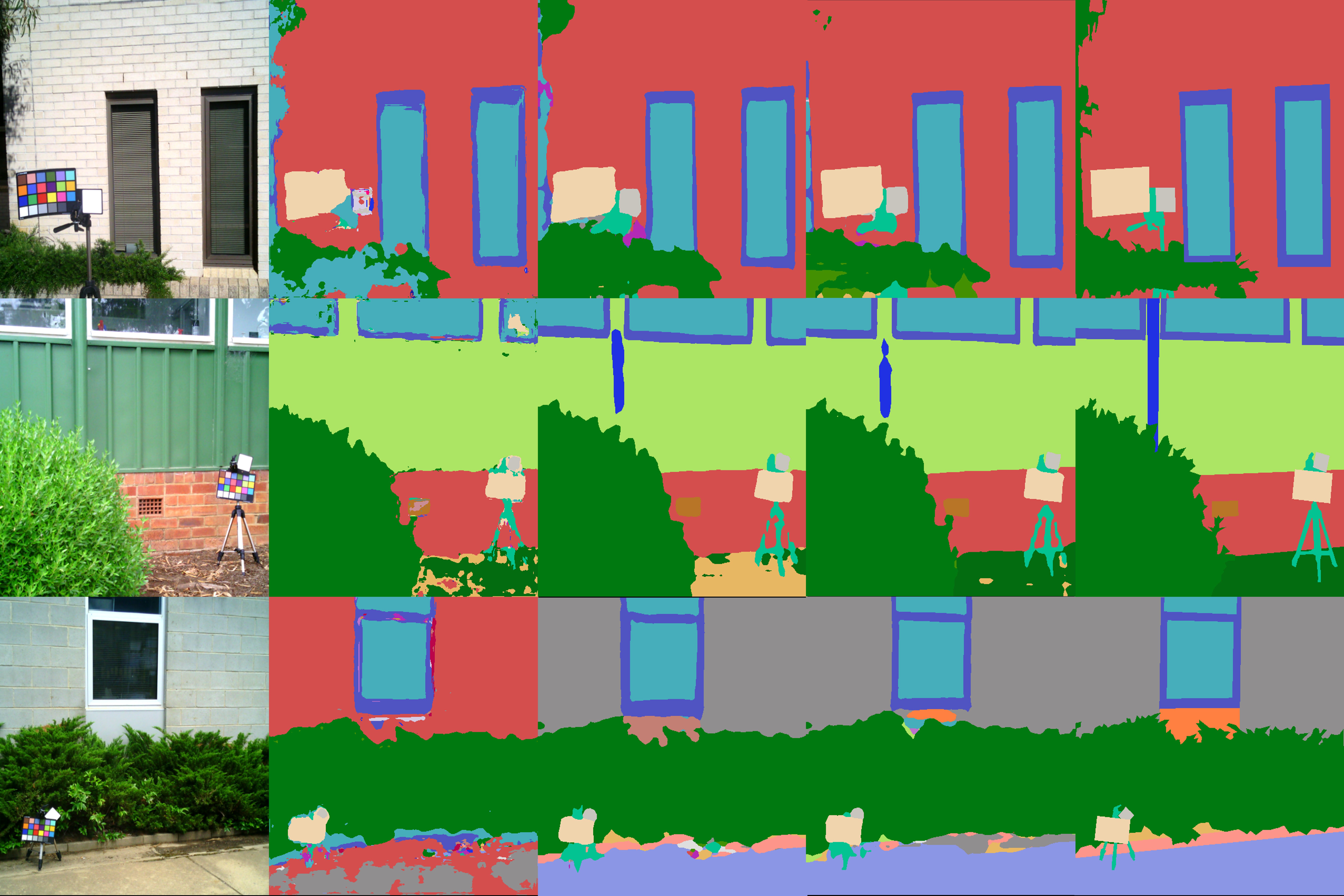}
\caption{Examples of semantic segmentation of the building materials using FCN and U-Net, from left to right panels; input RGB image, output of U-Net using RGB, output of FCN using RGB, output of FCN using RGB and HSI and the ground truth labels}
\label{fig:qualitative_results}
\end{figure}

The quantitative results of deep learning semantic segmentation algorithms on LIB-HSI dataset are presented in Table~\ref{tab:comparison}. The qualitative results are demonstrated in Figure~\ref{fig:qualitative_results}.
Using only RGB images to classify building materials from an image showed lower performance in comparison to using hyperspectral images or using both hyperspectral images and RGB. This could be due to the rich information captured in different wavelengths in hyperspectral data. Adding RGB images to the input did marginally improve the results, which might be attributed to the fact that hyperspectral images already include some pixel information contained in RGB images. 
\begin{table*}[!h]
\begin{center}
\caption{Semantic segmentation results using FCN and U-Net with ResNet50 backbone for classification of building materials from RGB, hyperspectral (HS).}
  \label{tab:comparison}

\begin{tabularx}{0.95\textwidth}{@{}lXXXXXX@{}}
\toprule
 &  &  & &\multicolumn{3}{c}{Average class} \\
\cline{5-7}
Modality & Method& Accuracy & IoU score & precision &  recall & IoU \\ \midrule

RGB       & FCN & 0.829 & 0.387 & 0.619 & 0.636 & 0.443\\
RGB       & U-Net & 0.687 & 0.162 & 0.360 & 0.426 &  0.236\\
\midrule
HS       & FCN &0.868  & 0.544 & 0.704 & 0.691 & 0.547\\
HS       & U-Net   & 0.750  & 0.180  & 0.465  &      0.510  &      0.318 \\
\midrule
RGB+HS      & FCN & 0.875  &  0.675  & 0.701  & 0.690 & 0.544 \\
\bottomrule
\end{tabularx}
  
\end{center}
\end{table*}
Inferred results from FCN as a patch-based method indicated a relatively higher performance in comparison to U-Net and therefore for the combination of RGB and hyperspectral images we only reported the FCN results. More advanced deep learning models, including light-weight models are required to be investigated on LIB-HSI.  

While using feature reduction algorithms, such as principal component analysis, is common in HSI  segmentation~\cite{zhao2021combination,zhang2021hyperspectral}, we decided not to use them as they could potentially increase the complexity while only offering a marginal improvement. In the future, we will investigate more advanced deep learning HSI dimension reduction methods. 

\section{Conclusion}
\label{cha:conc}
In this paper we introduced a novel HSI and RGB building facade dataset, that can be accessed at \href{https://doi.org/10.25919/vzwp-0w88}{\textcolor{blue}{CSIRO data access portal}}. Our LIB-HSI dataset was collected from a light industrial environment and covers a wide variety of building materials. We applied well established neural network algorithms on this dataset to segment the building materials. The results show that using both RGB and hyperspectral data can increase the performance of building material classification/segmentation. We envision that the LIB-HSI dataset can open up new research avenues for building material classification/segmentation.


%
\bibliographystyle{splncs04}
\bibliography{egbib}
\end{document}